\pdfoutput=1

\documentclass[11pt]{article}

\usepackage[]{acl}

\usepackage{times}
\usepackage{latexsym}
\usepackage{amsmath, amsfonts, amssymb}
\usepackage{pifont}     

\newcommand{\cmark}{\ding{51}} 
\newcommand{\xmark}{-} 

\usepackage[T1]{fontenc}

\usepackage[utf8]{inputenc}
\usepackage{booktabs}

\usepackage{microtype}

\usepackage{inconsolata}

\usepackage{graphicx}

\title{Visual-Aware Speech Recognition for Noisy Scenarios}

\author{Lakshmipathi Balaji \\
  IIIT-Hyderabad, India \\
  \texttt{lakshmipathi.balaji@research.iiit.ac.in} \\\And
  Karan Singla \\
  Whissle Inc., USA \\
  \texttt{ksingla@whissle.ai} \\}

\begin{document}
\maketitle
\begin{abstract}
Humans have the ability to utilize visual cues, such as lip movements and visual scenes, to enhance auditory perception, particularly in noisy environments. However, current Automatic Speech Recognition (ASR) or Audio-Visual Speech Recognition (AVSR) models often struggle in noisy scenarios. To solve this task, we propose a model that improves transcription by correlating noise sources to visual cues. Unlike works that rely on lip motion and require the speaker's visibility, we exploit broader visual information from the environment. This allows our model to naturally filter speech from noise and improve transcription, much like humans do in noisy scenarios. Our method re-purposes pretrained speech and visual encoders, linking them with multi-headed attention. This approach enables the transcription of speech and the prediction of noise labels in video inputs. We introduce a scalable pipeline to develop audio-visual datasets, where visual cues correlate to noise in the audio. We show significant improvements over existing audio-only models in noisy scenarios. Results also highlight that visual cues play a vital role in improved transcription accuracy.
\end{abstract}

\section{Introduction}
Automatic Speech Recognition (ASR) models have applications in many voice-enabled applications, including audio-video calls, intelligent virtual assistants, and media processing. These models are expected to work well in noisy conditions for their effective use in real-world scenarios. Several studies demonstrate that the human brain uses both audio and visual streams (e.g. lip motion, visual scenes) for listening, particularly when the speech is noisy ~\cite{vis_for_brain1, vis_for_brain2_lips, vis_for_brain3}. These models have applications where the visual stream is also available as additional input. These observations have led to the development of audio-visual speech recognition (AVSR) models.

Several AVSR models show that transcription can be improved in the noisy scenario by attending to lip-region movement~\cite{lips_noise_wacv24, av_hubert} and exploiting the correlation of visual scenes with spoken content~\cite{avatar_cvpr_23}. Recently~\citet{avsr_with_background_icassp24} show that background scenes can help in improving transcription in a given environment. However, its dependence on a manually collected dataset and limited alignment between visual context and audio hinder its scalability and effective utilization of visual cues.

Building on these insights, we address these limitations by proposing a scalable data creation pipeline and finetuning method that utilizes pretrained checkpoints. Our automated pipeline allows the mixing of audio-visual noise datasets with clean speech at variable noise ratios, eliminating the need for specialized datasets. In this work, we propose an architecture that integrates pretrained audio and visual encoders via Multi-Headed Attention. We hypothesize that training AVSR models with visual cues of the noise sources will improve speech recognition in noisy scenarios.

We use AudioSet~\cite{audioset} mixed with a clean speech corpus, People speech~\cite{peoples_speech} for finetuning purposes. We extract speech embeddings for each time-step in audio and then calculate enhanced representations by attending to visual features obtained from CLIP visual encoder~\cite{clip}. Our model takes (audio, video) pairs and finetunes the speech encoder for multi-modal speech recognition and noise label prediction jointly using CTC loss~\cite{ctc}. We hypothesize that leveraging the correlation between noise sources and visual cues will lead to more accurate transcription by providing richer context than background scene awareness alone. 

The resultant finetuned model improves transcription quality while also predicting noise labels. Ablation experiments further suggest that these improvements in transcription accuracy, are primarily due to our model's ability to attend to visual cues. The main contributions of this work are two-fold, (i) We propose a scalable dataset creation pipeline to develop audio-visual datasets, where visual cues correlate to noise sources in the audio. (ii) This work introduces a finetuning method that is visually aware of the noise while doing transcription.

\section{Related Work}


\subsection{Audio only noisy speech recognition}

Noise can be removed as a pre-processing step before being fed to ASR systems for improved transcription. Noise removal can be done either via signal enhancement techniques~\cite{signal_enhancement} and via source separation methods~\cite{demucs1, demucs2, hyperunmix}. 
Recent state-of-the-art E2E ASR systems enhance robustness in noisy environments by adding synthetic noise into their training datasets~\cite{wav2vec2, whisper, citrinet, wavlm}. However, purely audio-based models still face difficulties in extreme noise conditions, highlighting the need for multi-modal approaches, such as AVSR, which leverage visual cues to handle noise better.

\subsection{Audio-visual Speech Recognition}
\label{AVSR}
Recent studies propose AVSR models capable of exploiting visual cues for improved performance. Multiple works have focused on exploiting lip motion as additional information along with audio to improve transcription~\cite{av_hubert, lip_relate_1, lips_noise_wacv24}. In the context of full frame features, some works show that having visual cues related to the topics spoken helps with better word disambiguation~\cite{avatar_is22,avatar_cvpr_23}. However these works only see visual information to correlate with actual spoken content, instead, we focus on exploiting visual context as a cognition enhancer for ASR systems.
\section{Dataset Creation Pipeline}
\label{sec:dataset}
We aim to create a dataset where audio noise is closely correlated with the video content and each noise instance is uniquely annotated along ground truth transcriptions. To facilitate this, we have developed a dataset creation pipeline that selectively filters AudioSet~\cite{audioset} for videos and corresponding noise audio with annotated labels. We then mix noise-labeled videos with the People's Speech dataset~\cite{peoples_speech}, that have ground-truth transcriptions. Further details are discussed below.
\subsection{Filtering AudioSet}
\textit{AudioSet}~\cite{audioset} comprises of 2 million human-labelled, 10-second audio clips from YouTube, categorized into 632 audio event classes arranged hierarchically. This work targets only the videos associated with a noise label; thus, we exclude any video labelled with speech or human voice. We limit our scope to videos that only have a single noise label. We found that there is a big skew in the class distribution of noise labels, therefore we only select labels having at least 750 samples. This filtered subset of AudioSet has 44 unique noise labels (e.g. car, water, fireworks).


\subsection{Mixing with People's Speech}
\textit{People's Speech}~\cite{peoples_speech} is an ASR dataset featuring 30K hours of transcribed English speech from a diverse range of speakers. We utilize clean subset of it for our dataset. Since AudioSet videos are of 10 seconds each, we select speech samples longer than 10 seconds and then trim both audio and transcripts.
We take a clean speech sample and run an off-the-shelf forced aligner from the NeMo toolkit~\cite{nemo}. The forced-aligned output provides word time stamps, allowing us to trim both audio and transcripts to a 10-second duration. We append the noise label as the final word to the transcripts, enabling the model to learn both transcription and noise label prediction for each sample.

We process our filtered AudioSet (10-second video clips) and clean speech recordings to generate samples consisting of: video (without audio), corresponding noisy audio, clean speech, and corresponding transcripts. A noisy speech is obtained by mixing the clean speech recording with the original noisy audio extracted from the same video clip in a one-to-one correspondence.

Finally, we divide the dataset curated into training, validation, and testing subsets, ensuring each set contains a uniform distribution of noise samples from AudioSet. We refer to this dataset as the Visual-Aware Noisy Speech (VANS) dataset in further sections. The current VANS dataset contains 28K samples, providing 75 hours of training data, and 2K samples each contributing 6.1 hours for validation and testing. It is important to note that this dataset is scalable and can be expanded by incorporating more samples from AudioSet that may contain multiple labels, as well as more samples from People's Speech. Furthermore, we can enhance the dataset by dynamically altering the sample mixing mappings during model training to create augmentations.

\section{Method}


\subsection{Architecture}
\label{architecture}
\begin{figure}
    \centering
    \includegraphics[width=1\linewidth]{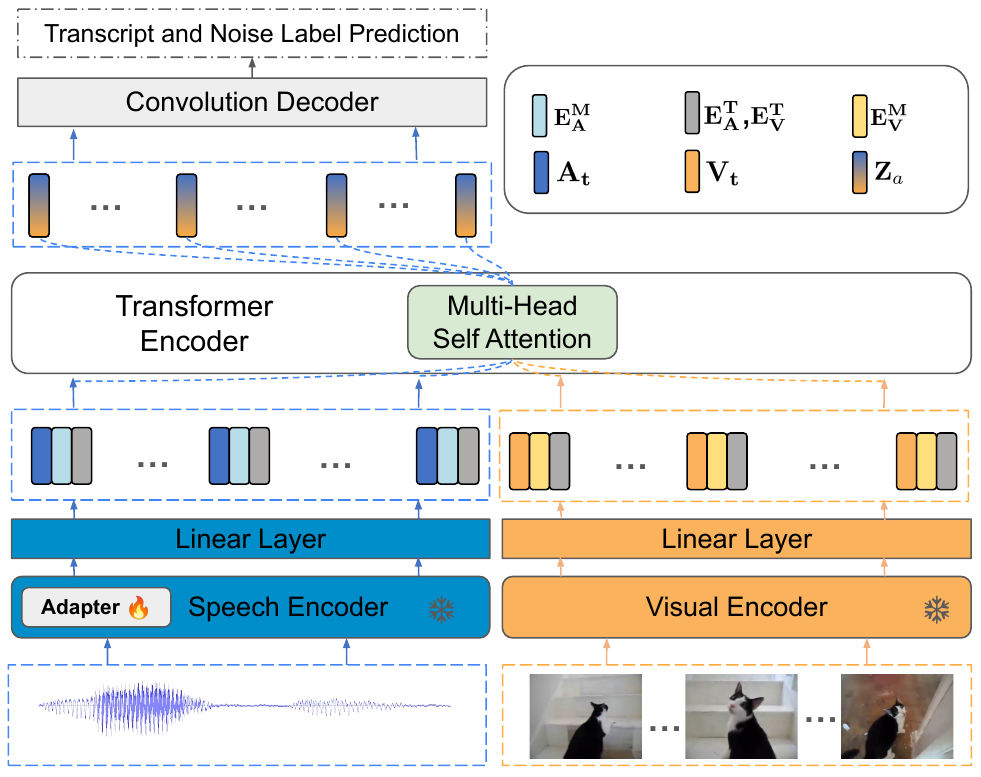}
    \caption{A visualization of our architecture. Speech and Visual representations are first obtained from their respective encoders, then aligned and enhanced via a Transformer-based Multi-Head Self-Attention mechanism. The output is then decoded using a convolutional decoder for simultaneous transcript and noise label prediction.}
    \label{fig:architecture}
\end{figure}

Drawing inspiration from previous works~\cite{avatar_is22, lips_noise_wacv24}, we employ a late fusion strategy. We re-purpose the encoder from a pretrained E2E ASR\footnote{\url{https://catalog.ngc.nvidia.com/orgs/nvidia/teams/nemo/models/stt_en_conformer_ctc_large}}, based on Conformer architecture \cite{conformer}. For an input noisy audio, we get $\mathbf{H_a}$ that represents audio embeddings from the speech encoder. Similarly, we use CLIP's ViT-L/14 image encoder~\cite{clip} to extract visual features $\mathbf{H_v}$. Note that, both encoders remain frozen; however, to enhance learning from noisy speech, we finetune the speech encoder using adapters. A visual overview of this approach is presented in Figure \ref{fig:architecture}. $\mathbf{H_a}$ and $\mathbf{H_v}$ are then brought to common dimensionality using dense layers $\mathbf{W_A}$ and $\mathbf{W_V}$ to get $\mathbf{A_t}$ and $\mathbf{V_t}$ respectively. Formally,

\begin{equation}
\label{eq:audio_eq}
\overline{\mathbf{A}_t} = \mathbf{W_A} \mathbf{H_a} + \mathbf{E^M_A} + \mathbf{E^T_A},
\end{equation}
\begin{equation}
\label{eq:video_eq}
\overline{\mathbf{V}_t} = \mathbf{W_V} \mathbf{H_v} + \mathbf{E^M_V} + \mathbf{E^T_V}.
\end{equation}

$\mathbf{E^T_A}$ and $\mathbf{E^T_V}$ represent the positional embeddings for the audio and video time series, respectively. We use separate positional embeddings for audio and visual features to enhance the system's ability to track context across both modalities. Additionally, $\mathbf{E^M_A}$ for audio and $\mathbf{E^M_V}$ for video are modality embeddings, enabling the system to effectively distinguish between audio and visual information.

$\overline{\mathbf{A_t}}$ and $\overline{\mathbf{V_t}}$ from \eqref{eq:audio_eq} and \eqref{eq:video_eq} are then passed through a standard transformer encoder block, facilitating Multi-Head Self-Attention across the modalities~\cite{transformer}. This cross-modal interaction yields outputs $\mathbf{Z}_a$ for audio and $\mathbf{Z}_v$ for video respectively. For our task, we only utilize the visual-aware audio outputs $\mathbf{Z}_a$ and ignore $\mathbf{Z}_v$. $\mathbf{Z}_a$ is then processed through a convolutional decoder and then optimized for transcription task using standardized CTC loss. In our case, the last word in the transcripts refers to the noise label. 



\subsection{Base Model Pretraining}
\label{pretraining}

Existing ASR models and tokenizers typically include only transcription-related tokens, whereas our model requires the final token to represent noise label, which is not covered by the pretrained E2E ASR tokenizer. Following \cite{1step}, we extended the tokenizer to include special tokens for noise labels, necessitating the reinitialization of the prediction layer in the convolutional decoder. To adapt the model, we performed additional pretraining of ASR with this extended tokenizer using 420 hours of People's Speech data and CTC loss. This produced a pretrained speech encoder capable of handling transcription tokens along with noise label tokens that are predicted as the last word.

\section{Experiments \& Results}
\textbf{Implementation details}. 
Our experiments utilize a pretrained model, initially trained solely on transcription task without visual inputs, as described earlier. For visual information, we extract CLIP features at 5 fps. We use a Transformer Encoder with 4 layers with a dimensionality of 512.
We assess model performance using Word Error Rate (WER) for transcription task and noise label prediction accuracy. For each prediction, we first strip away the noise label at the end, if present, and then compare the remaining transcript against the ground truth transcript of the audio clip. We use the extracted noise label to evaluate the accuracy of the noise label prediction task.

\textbf{Models}. We conducted a series of experiments to demonstrate the improved performance of our model in noisy conditions by leveraging visual information. Thus, we selected 10dB SNR noisy speech samples for our experiments and train audio and audio-visual models. We recognize that it is impractical to train a separate model for each possible noise level, therefore we adopt a uniform sampling strategy to dynamically choose the SNR values in the range of -5 dB to +5 dB for each sample. This method, termed AV-UNI-SNR, ensures that our model encounters a varied but controlled set of noise scenarios, thus enhancing its ability to generalize across similar conditions.





\subsection{Results}
\label{subsec:results}


\begin{table}[h!]
    \centering
    \resizebox{\columnwidth}{!}{%
    \begin{tabular}{@{}clcccccr@{}}
    \toprule
    \hline
    & Model & SNR (dB) & $P_r$ & $V_T$ & $V_I$ & WER & ACC (\%)  \\ \midrule
    \hline
    1 & Conformer-CTC   & - & \xmark & \xmark & \xmark & 26.99 & -  \\
    \hline
    2 & A-SNR   & 10 & \cmark & \xmark & \xmark & 23.30 & 2.98  \\
    3 & A-UNI-SNR   & [-5,5] & \cmark & \xmark & \xmark & 23.11 & 4.54 \\
    \hline
    3 & AV-SNR & 10 & \cmark & \cmark & \cmark & 21.83 & \textbf{60.95}  \\
    4 & AV-SNR  & 10 & \xmark & \cmark & \cmark & 23.59 & 58.59  \\
    \hline
    5 & AV-UNI-SNR & [-5,5] & \cmark & \cmark & \cmark & \textbf{20.71} & 54.23  \\
    6 & AV-UNI-SNR  & [-5,5] & \cmark & \cmark & \xmark & 22.29 & 2.36  \\
    \bottomrule
    \hline
    \end{tabular}%
    }
    \caption{Model Performance at SNR 10 dB. $P_r$ refers to pretraining, $V_T$ refers to visual information available during training, and $V_I$ refers to visual information available during inference. "A" indicates models using only audio, while "AV" represents models utilizing both audio and video while training. "UNI" refers to models trained with uniformly sampled SNR levels. For details, please refer to section \ref{subsec:results}.}

    \label{tab:model_performance}
\end{table}

\vspace{-2pt}
Table \ref{tab:model_performance} presents the results of our experiments. On comparing R2 and R4 shows gains over the audio-only model in transcription accuracy with visual awareness. Notably, results depict a big gain in the correct prediction of noise labels when model learns to exploit cues from visual background. This proves our hypothesis that the correlation of noise with the visual cues helps with improved transcription and noise label predictions. The comparison between R4 and R5 shows the importance of pretraining, in preparing the model for both transcription and noise prediction tasks.

Results for AV-UNI-SNR models show the best performance overall. Performance gains are higher when visual information is provided at both finetuning and inference time. However, results in the last row show our model improves over the audio-only model (R3) even when visual information is not provided at inference time. This suggests that models trained with visual guidance for noise detection also perform well when only audio is used during inference. It shows that models trained with visual cues develop a more nuanced understanding of complex acoustic environments than audio-only models. However, it falls short in predicting noise labels without visual input. The model naturally tends to rely on
video context for noise prediction, as it offers clearer cues. Consequently, when tested with only audio inputs, the model’s performance on the noise prediction task declines. We discuss more about results across SNRs and computational costs in section \ref{Appendix}.



\begin{table}[h!]
\centering
\renewcommand{\arraystretch}{1.2} 
\resizebox{\columnwidth}{!}{ 
\begin{tabular}{@{}clcc@{}}
\hline
 & \textbf{Models}      & \textbf{LS test-clean} & \textbf{LS test-other} \\ \hline
1 & Conformer-CTC~\cite{conformer}          & 31.07                 & 39.89                 \\ 
2 & A-UNI-SNR \textit{(Ours)}    & 28.05                 & 37.91                 \\ 
3 & AV-UNI-SNR \textit{(Ours)}      & \textbf{27.86}             & \textbf{37.47}                  \\ \hline
\end{tabular}
}
\caption{Models Performance at SNR 0 dB \\ on LibriSpeech (LS) Test Sets.}
\label{tab:ls_results}
\end{table}

\vspace{-2pt}
\textbf{Out-of-Domain Evaluation.} While AV-UNI-SNR is pretrained on People's Speech, and Conformer-CTC is pretrained on a broader range of datasets including People's Speech and LibriSpeech~\cite{librispeech}, there may be concerns that AV-UNI-SNR's superior performance on noisy audio is due to its specialized training on People's Speech. 
To address this, we conducted an additional experiment using LibriSpeech, mixed with AudioSet samples as described in section \ref{sec:dataset}. Importantly, LibriSpeech is within the domain for Conformer-CTC but out-of-domain for our model. As shown in Table \ref{tab:ls_results}, our model still outperforms R1 and R2 on this dataset as well, confirming that R3 is robust in noisy environments even with out-of-domain data.

\section{Conclusion}
In this work, we show that exploiting visual cues with audio signals significantly improves transcription accuracy for noisy scenarios. Our automated dataset creation pipeline, designed to align noise with visual cues, provides a promising foundation for enhancing AVSR models. We show that models trained across varied SNR levels, especially the AV-UNI-SNR model, excel in diverse noise conditions. Our proposed method is easily adaptable to other pretrained architectures and checkpoints.

\section*{Limitations}
While AudioSet provides a scalable foundation, the success of this approach relies heavily on its fine-grained noise-to-video correlations. These annotations, although extensive, are still manually curated and may not fully capture the complexity of real-world noisy environments. Incorporating visual inputs during inference introduces computational overhead, primarily due to the use of a pretrained CLIP visual encoder. While this overhead exists for achieving the best performance, our approach mitigates this by outperforming audio-only models even when used with only audio inputs during inference. However, for scenarios demanding the highest accuracy, the additional computational cost remains a trade-off.

\bibliography{custom}

\begin{thebibliography}{26}
\providecommand{\natexlab}[1]{#1}

\bibitem[{Baevski et~al.(2020)Baevski, Zhou, rahman Mohamed, and Auli}]{wav2vec2}
Alexei Baevski, Henry Zhou, Abdel rahman Mohamed, and Michael Auli. 2020.
\newblock \href {https://api.semanticscholar.org/CorpusID:219966759} {wav2vec 2.0: A framework for self-supervised learning of speech representations}.
\newblock \emph{ArXiv}, abs/2006.11477.

\bibitem[{Boots et~al.(2020)Boots, Mead, Rawashdeh, Bellapart, Townsend, Paratz, Garner, Clement, Oddy, and behalf of the Circadian Investigators~in Critical~Illness}]{vis_for_brain3}
Robert~James Boots, Gabrielle Mead, Oliver Rawashdeh, Judith Bellapart, Shane Townsend, Jennifer~D. Paratz, Nicholas Garner, Pierre Clement, David Oddy, and On~behalf of the Circadian Investigators~in Critical~Illness. 2020.
\newblock \href {https://api.semanticscholar.org/CorpusID:238118950} {Circadian hygiene in the icu environment (chie) study}.
\newblock \emph{Critical Care and Resuscitation}, 22:361 -- 369.

\bibitem[{Burchi and Timofte(2023)}]{lips_noise_wacv24}
Maxime Burchi and Radu Timofte. 2023.
\newblock \href {https://api.semanticscholar.org/CorpusID:255416090} {Audio-visual efficient conformer for robust speech recognition}.
\newblock \emph{2023 IEEE/CVF Winter Conference on Applications of Computer Vision (WACV)}, pages 2257--2266.

\bibitem[{Chen et~al.(2021)Chen, Wang, Chen, Wu, Liu, Chen, Li, Kanda, Yoshioka, Xiao, Wu, Zhou, Ren, Qian, Qian, Zeng, and Wei}]{wavlm}
Sanyuan Chen, Chengyi Wang, Zhengyang Chen, Yu~Wu, Shujie Liu, Zhuo Chen, Jinyu Li, Naoyuki Kanda, Takuya Yoshioka, Xiong Xiao, Jian Wu, Long Zhou, Shuo Ren, Yanmin Qian, Yao Qian, Micheal Zeng, and Furu Wei. 2021.
\newblock \href {https://api.semanticscholar.org/CorpusID:239885872} {Wavlm: Large-scale self-supervised pre-training for full stack speech processing}.
\newblock \emph{IEEE Journal of Selected Topics in Signal Processing}, 16:1505--1518.

\bibitem[{D{\'e}fossez(2021)}]{demucs2}
Alexandre D{\'e}fossez. 2021.
\newblock Hybrid spectrogram and waveform source separation.
\newblock In \emph{Proceedings of the ISMIR 2021 Workshop on Music Source Separation}.

\bibitem[{Gabeur et~al.(2022)Gabeur, Seo, Nagrani, Sun, Karteek, and Schmid}]{avatar_is22}
Valentin Gabeur, Paul~Hongsuck Seo, Arsha Nagrani, Chen Sun, Alahari Karteek, and Cordelia Schmid. 2022.
\newblock \href {https://api.semanticscholar.org/CorpusID:249674615} {Avatar: Unconstrained audiovisual speech recognition}.
\newblock In \emph{Interspeech}.

\bibitem[{Galvez et~al.(2021)Galvez, Diamos, Ciro, Cer'on, Achorn, Gopi, Kanter, Lam, Mazumder, and Reddi}]{peoples_speech}
Daniel Galvez, Gregory~Frederick Diamos, Juan Ciro, Juan~Felipe Cer'on, Keith Achorn, Anjali Gopi, David Kanter, Maximilian Lam, Mark Mazumder, and Vijay~Janapa Reddi. 2021.
\newblock \href {https://api.semanticscholar.org/CorpusID:237265975} {The people's speech: A large-scale diverse english speech recognition dataset for commercial usage}.
\newblock \emph{ArXiv}, abs/2111.09344.

\bibitem[{Gemmeke et~al.(2017)Gemmeke, Ellis, Freedman, Jansen, Lawrence, Moore, Plakal, and Ritter}]{audioset}
Jort~F. Gemmeke, Daniel P.~W. Ellis, Dylan Freedman, Aren Jansen, Wade Lawrence, R.~Channing Moore, Manoj Plakal, and Marvin Ritter. 2017.
\newblock \href {https://doi.org/10.1109/ICASSP.2017.7952261} {Audio set: An ontology and human-labeled dataset for audio events}.
\newblock In \emph{2017 IEEE International Conference on Acoustics, Speech and Signal Processing (ICASSP)}, pages 776--780.

\bibitem[{Graves and Graves(2012)}]{ctc}
Alex Graves and Alex Graves. 2012.
\newblock Connectionist temporal classification.
\newblock \emph{Supervised sequence labelling with recurrent neural networks}, pages 61--93.

\bibitem[{Gulati et~al.(2020)Gulati, Qin, Chiu, Parmar, Zhang, Yu, Han, Wang, Zhang, Wu, and Pang}]{conformer}
Anmol Gulati, James Qin, Chung-Cheng Chiu, Niki Parmar, Yu~Zhang, Jiahui Yu, Wei Han, Shibo Wang, Zhengdong Zhang, Yonghui Wu, and Ruoming Pang. 2020.
\newblock \href {https://api.semanticscholar.org/CorpusID:218674528} {Conformer: Convolution-augmented transformer for speech recognition}.
\newblock \emph{ArXiv}, abs/2005.08100.

\bibitem[{Huang and Kingsbury(2013)}]{lip_relate_1}
Jing Huang and Brian Kingsbury. 2013.
\newblock \href {https://doi.org/10.1109/ICASSP.2013.6639140} {Audio-visual deep learning for noise robust speech recognition}.
\newblock In \emph{2013 IEEE International Conference on Acoustics, Speech and Signal Processing}, pages 7596--7599.

\bibitem[{Karan et~al.(2023)Karan, Shahab, Yeon-Jun, Andrej, Moreno, Srinivas, and Benjamin}]{1step}
Singla Karan, Jalalv Shahab, Kim Yeon-Jun, Ljolje Andrej, Daniel~Antonio Moreno, Bangalore Srinivas, and Stern Benjamin. 2023.
\newblock \href {https://api.semanticscholar.org/CorpusID:271162394} {1-step speech understanding and transcription using ctc loss}.
\newblock In \emph{ICON}.

\bibitem[{Kuchaiev et~al.(2019)Kuchaiev, Li, Nguyen, Hrinchuk, Leary, Ginsburg, Kriman, Beliaev, Lavrukhin, Cook, Castonguay, Popova, Huang, and Cohen}]{nemo}
Oleksii Kuchaiev, Jason Li, Huyen Nguyen, Oleksii Hrinchuk, Ryan Leary, Boris Ginsburg, Samuel Kriman, Stanislav Beliaev, Vitaly Lavrukhin, Jack Cook, Patrice Castonguay, Mariya Popova, Jocelyn Huang, and Jonathan~M. Cohen. 2019.
\newblock \href {https://api.semanticscholar.org/CorpusID:202712805} {Nemo: a toolkit for building ai applications using neural modules}.
\newblock \emph{ArXiv}, abs/1909.09577.

\bibitem[{Luo et~al.(2024)Luo, Liu, Sun, and Sun}]{avsr_with_background_icassp24}
Cheng Luo, Yiguang Liu, Wenhui Sun, and Zhoujian Sun. 2024.
\newblock \href {https://api.semanticscholar.org/CorpusID:268603096} {Multi-modality speech recognition driven by background visual scenes}.
\newblock \emph{ICASSP 2024 - 2024 IEEE International Conference on Acoustics, Speech and Signal Processing (ICASSP)}, pages 10926--10930.

\bibitem[{Majumdar et~al.(2021)Majumdar, Balam, Hrinchuk, Lavrukhin, Noroozi, and Ginsburg}]{citrinet}
Somshubra Majumdar, Jagadeesh Balam, Oleksii Hrinchuk, Vitaly Lavrukhin, Vahid Noroozi, and Boris Ginsburg. 2021.
\newblock Citrinet: Closing the gap between non-autoregressive and autoregressive end-to-end models for automatic speech recognition.
\newblock \emph{arXiv preprint arXiv:2104.01721}.

\bibitem[{McGurk and MacDonald(1976)}]{vis_for_brain2_lips}
Harry McGurk and John MacDonald. 1976.
\newblock \href {https://api.semanticscholar.org/CorpusID:4171157} {Hearing lips and seeing voices}.
\newblock \emph{Nature}, 264:746--748.

\bibitem[{Panayotov et~al.(2015)Panayotov, Chen, Povey, and Khudanpur}]{librispeech}
Vassil Panayotov, Guoguo Chen, Daniel Povey, and Sanjeev Khudanpur. 2015.
\newblock \href {https://doi.org/10.1109/ICASSP.2015.7178964} {Librispeech: An asr corpus based on public domain audio books}.
\newblock In \emph{2015 IEEE International Conference on Acoustics, Speech and Signal Processing (ICASSP)}, pages 5206--5210.

\bibitem[{Petermann et~al.(2023)Petermann, Wichern, Subramanian, and {Le Roux}}]{hyperunmix}
Darius Petermann, Gordon Wichern, Aswin Subramanian, and Jonathan {Le Roux}. 2023.
\newblock Hyperbolic audio source separation.
\newblock In \emph{Proc. IEEE International Conference on Acoustics, Speech and Signal Processing (ICASSP)}.

\bibitem[{Radford et~al.(2021)Radford, Kim, Hallacy, Ramesh, Goh, Agarwal, Sastry, Askell, Mishkin, Clark, Krueger, and Sutskever}]{clip}
Alec Radford, Jong~Wook Kim, Chris Hallacy, Aditya Ramesh, Gabriel Goh, Sandhini Agarwal, Girish Sastry, Amanda Askell, Pamela Mishkin, Jack Clark, Gretchen Krueger, and Ilya Sutskever. 2021.
\newblock \href {https://api.semanticscholar.org/CorpusID:231591445} {Learning transferable visual models from natural language supervision}.
\newblock In \emph{International Conference on Machine Learning}.

\bibitem[{Radford et~al.(2022)Radford, Kim, Xu, Brockman, McLeavey, and Sutskever}]{whisper}
Alec Radford, Jong~Wook Kim, Tao Xu, Greg Brockman, Christine McLeavey, and Ilya Sutskever. 2022.
\newblock \href {https://api.semanticscholar.org/CorpusID:252923993} {Robust speech recognition via large-scale weak supervision}.
\newblock \emph{ArXiv}, abs/2212.04356.

\bibitem[{Rouard et~al.(2023)Rouard, Massa, and D{\'e}fossez}]{demucs1}
Simon Rouard, Francisco Massa, and Alexandre D{\'e}fossez. 2023.
\newblock Hybrid transformers for music source separation.
\newblock In \emph{ICASSP 23}.

\bibitem[{Seo et~al.(2023)Seo, Nagrani, and Schmid}]{avatar_cvpr_23}
Paul~Hongsuck Seo, Arsha Nagrani, and Cordelia Schmid. 2023.
\newblock \href {https://doi.org/10.1109/CVPR52729.2023.02195} {Avformer: Injecting vision into frozen speech models for zero-shot av-asr}.
\newblock In \emph{2023 IEEE/CVF Conference on Computer Vision and Pattern Recognition (CVPR)}, pages 22922--22931.

\bibitem[{Shi et~al.(2022)Shi, Hsu, Lakhotia, and rahman Mohamed}]{av_hubert}
Bowen Shi, Wei-Ning Hsu, Kushal Lakhotia, and Abdel rahman Mohamed. 2022.
\newblock \href {https://api.semanticscholar.org/CorpusID:245769552} {Learning audio-visual speech representation by masked multimodal cluster prediction}.
\newblock \emph{ArXiv}, abs/2201.02184.

\bibitem[{Steinmetz et~al.(2023)Steinmetz, Walther, and Reiss}]{signal_enhancement}
Christian~J Steinmetz, Thomas Walther, and Joshua~D Reiss. 2023.
\newblock High-fidelity noise reduction with differentiable signal processing.
\newblock \emph{arXiv preprint arXiv:2310.11364}.

\bibitem[{Sumby and Pollack(1954)}]{vis_for_brain1}
William~H. Sumby and Irwin Pollack. 1954.
\newblock \href {https://api.semanticscholar.org/CorpusID:121993592} {Visual contribution to speech intelligibility in noise}.
\newblock \emph{Journal of the Acoustical Society of America}, 26:212--215.

\bibitem[{Vaswani(2017)}]{transformer}
A~Vaswani. 2017.
\newblock Attention is all you need.
\newblock \emph{Advances in Neural Information Processing Systems}.

\end{thebibliography}
\appendix
\clearpage
\section{Appendix}
\label{Appendix}
In this section, we present additional experiments conducted across various SNRs (\ref{subsec:App-Results_across_SRNs}), analyze the computational costs of our AV model in (\ref{subsec:App-Comp_Costs}) and discuss the future works \ref{subsec:App-Future_works}.
\subsection{Results across SNRs}
\label{subsec:App-Results_across_SRNs}

\begin{figure}[ht] 
    \centering
    \includegraphics[width=1\linewidth]{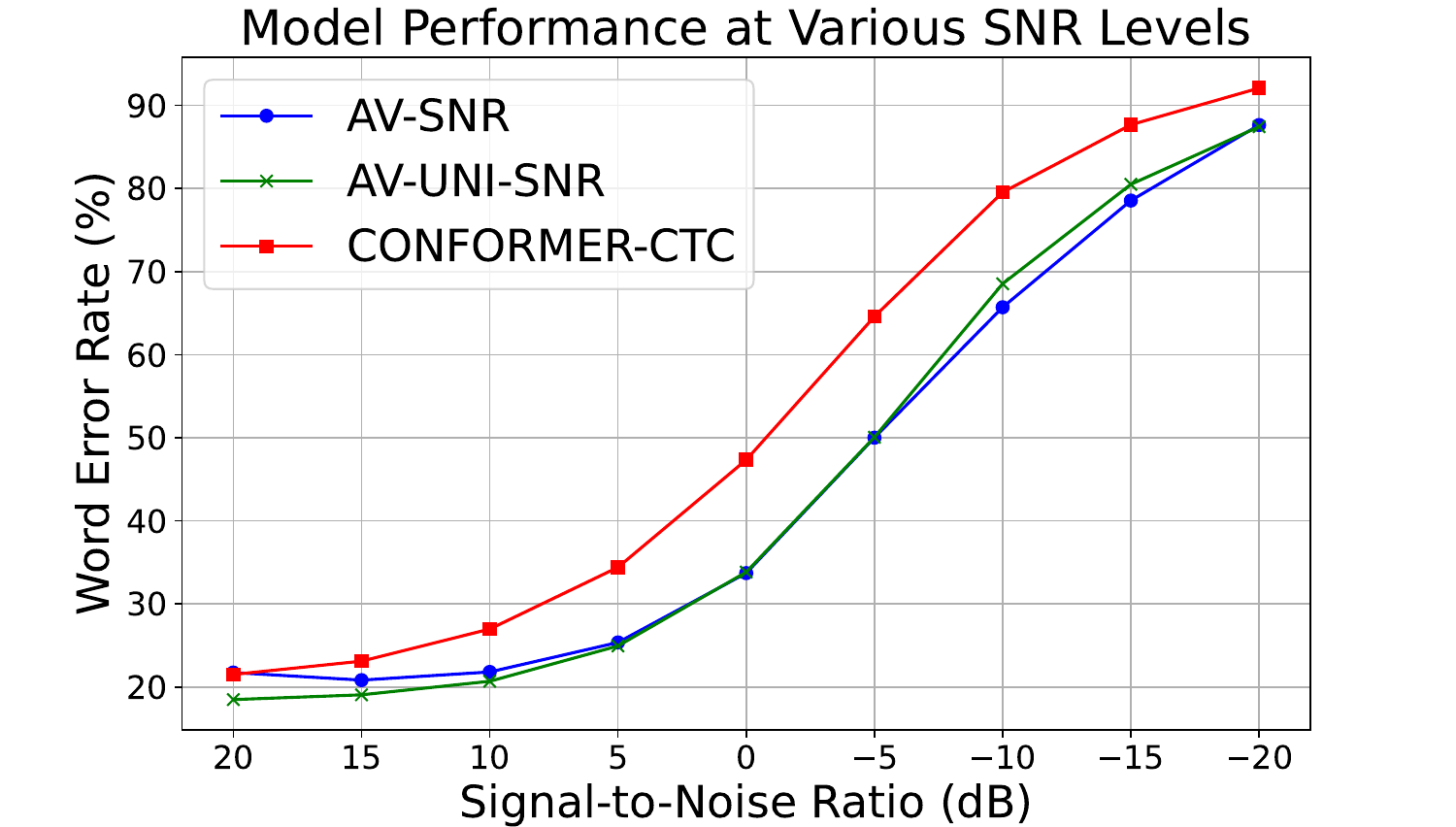}
    \caption{Model performance comparison across SNR levels on test set of proposed VANS dataset, highlighting AV-UNI-SNR's robustness in lower SNR environments.}
    \label{fig:results_plot1}
\end{figure}
    
The results in Figure \ref{fig:results_plot1} show that AV-UNI-SNR generalizes well across varying SNR levels, outperforming the individual models in lower SNR conditions (below -5 dB). However, models trained at fixed SNRs perform better at higher SNR values. These findings, along with the results from Table \ref{tab:model_performance}, suggest that training on variable SNR values, as in the AV-UNI-SNR model, enables robust performance across noisy conditions,
and using visual cues further enhances generalization, even when visual cues are absent during inference.

\textbf{Training Details.} Our AVSR model was trained for 10 epochs on a single L40S GPU with a batch size of 96, completing in approximately 8 hours. The model employs a 4-layer Transformer Encoder with 8 attention heads and a dimensionality of 512. Linear adapters with a dimensionality of 64 are incorporated into the speech encoder. For all other hyperparameters, we adhere to the NEMO toolkit defaults.

\subsection{Computational costs?}
\label{subsec:App-Comp_Costs}
\begin{table}[ht]
\centering
\renewcommand{\arraystretch}{1.2} 
\resizebox{\columnwidth}{!}{ 
\begin{tabular}{@{}lccccc@{}}
\hline
 & \textbf{Models}                                 & \textbf{Params} & \textbf{A} & \textbf{V} & \textbf{WER} \\ \hline
1 & Conformer-CTC Large        & 120M            & \cmark & - & 26.99      \\ 
2 & Conformer-CTC XLarge (XL)        & 635M            & \cmark & - & 26.15      \\ \hline
3 & A-UNI-SNR \textit{(Large Backbone)} & 150M            & \cmark & - & 23.11      \\
4 & A-UNI-SNR \textit{(XL Backbone)} & 665M            & \cmark & - & 22.34      \\ \hline
5 & AV-UNI-SNR \textit{(Ours)}    & 453M            & \cmark & \cmark & 20.71      \\ 
6 & AV-UNI-SNR \textit{(Ours)}    & 150M            & \cmark & - & 22.29      \\ \hline
\end{tabular}
}
\caption{Comparison of Models, Parameters, Modalities, and WER on Test Set of proposed dataset at 10dB.}
\label{tab:comparison_models}
\end{table}

We discuss the computational costs of our AV model in Table~\ref{tab:comparison_models}. Using visual inputs at inference requires an additional 300M parameters for CLIP feature extraction R5, increasing computational overhead compared to audio-only models. However, our AV-UNI-SNR model is flexible, supporting both audio-visual and audio-only inference. Notably, when used with only audio R6 it requires just 30M more parameters than the Conformer-CTC Large model (R1). Despite this smaller increase in parameters, our AV-UNI-SNR model outperforms the A-UNI-SNR XL model (R4), trained on audio-only data with 4x more parameters, demonstrating the superior efficiency and performance of our AV framework.

\subsection{Future Work}
\label{subsec:App-Future_works}
We plan to improve our model by exploring additional pretrained speech and visual encoder checkpoints and expanding our dataset pipeline to include AudioSet samples with multiple noise labels, enhancing visual context awareness. Furthermore, we plan to extend this approach to scalable audio-visual speech transcription, incorporating not only noise labels but also other visual cues and related events as tags.

Our framework discussed in section \ref{sec:dataset} has the potential to scale up and generate over 4000 hours of data by leveraging the full clean subset of People’s Speech and AudioSet. This scalability enables the community to adopt and expand our approach for AVSR training, facilitating the development of models that leverage our AV training strategy. Such models could achieve superior performance with audio-only inputs at test time compared to those trained solely with audio.
\end{document}